\pdfoutput=1

\documentclass{article}

\PassOptionsToPackage{numbers, compress}{natbib}

\usepackage[preprint]{neurips_2022}

\usepackage[utf8]{inputenc} 
\usepackage[T1]{fontenc}    
\usepackage{hyperref}       
\usepackage{url}            
\usepackage{booktabs}       
\usepackage{amsfonts}       
\usepackage{nicefrac}       
\usepackage{microtype}      
 
\usepackage{graphicx}
\usepackage{amsmath}
\usepackage{amssymb}
\usepackage{booktabs}

\usepackage{float}
\usepackage{mathtools}
\usepackage{framed}
\usepackage{pbox}
\usepackage{afterpage}
\usepackage{url}
\usepackage{array}
\usepackage{amsfonts}
\usepackage[table]{xcolor}

\usepackage{multirow}
\usepackage{booktabs}
\usepackage{relsize}
\usepackage{xspace}
\usepackage{caption}
\usepackage{subcaption}
\newcommand{\ignore}[1]{}

\newcommand{\norm}[1]{\left\lVert#1\right\rVert}

\usepackage{framed}	
\usepackage{url}
\usepackage{enumerate}
\usepackage{color}

\usepackage{changepage}

\usepackage{algpseudocode}
\usepackage{algorithm}

\usepackage{amsthm}
\theoremstyle{definition}
\theoremstyle{plain}

\newcommand{\SKIP}[1]{}

\newcommand{\bfzero}{\boldsymbol{0}}

\def\amin#1{{\underset{#1}{\operatorname{argmin}}}}%

\usepackage{xcolor}

\usepackage{bbm}

\graphicspath{{images/}}

\usepackage{amsmath,amsfonts,bm}

\def\figref#1{figure~\ref{#1}}

\def\twofigref#1#2{figures \ref{#1} and \ref{#2}}

\def\secref#1{section~\ref{#1}}

\def\eqref#1{equation~\ref{#1}}

\def\1{\bm{1}}

\def\rvh{{\mathbf{h}}}
\def\rvu{{\mathbf{i}}}

\def\rvu{{\mathbf{u}}}
\def\rvv{{\mathbf{v}}}

\def\rvx{{\mathbf{x}}}
\def\rvy{{\mathbf{y}}}
\def\rvz{{\mathbf{z}}}

\def\rmA{{\mathbf{A}}}

\def\mA{{\bm{A}}}

\DeclareMathAlphabet{\mathsfit}{\encodingdefault}{\sfdefault}{m}{sl}
\SetMathAlphabet{\mathsfit}{bold}{\encodingdefault}{\sfdefault}{bx}{n}

\def\gB{{\mathcal{B}}}

\def\gD{{\mathcal{D}}}

\def\gH{{\mathcal{H}}}

\def\gN{{\mathcal{N}}}

\def\gR{{\mathcal{R}}}

\def\gT{{\mathcal{T}}}

\def\gZ{{\mathcal{Z}}}

\newcommand{\R}{\mathbb{R}}

\usepackage[page]{appendix}
\usepackage{chngcntr}
\usepackage{etoolbox}

\AtBeginEnvironment{subappendices}{%
\section*{Appendix}
\addcontentsline{toc}{section}{Appendix}
}

\usepackage{xspace}
\makeatletter
\DeclareRobustCommand\onedot{\futurelet\@let@token\@onedot}
\def\@onedot{\ifx\@let@token.\else.\null\fi\xspace}

\def\eg{\emph{e.g}\onedot} 
\def\ie{\emph{i.e}\onedot}

\makeatother

\usepackage{enumitem}

\def\figref#1{Fig.~\ref{#1}}
\def\twofigref#1#2{Figs.~\ref{#1} and~\ref{#2}}
\def\secref#1{Sec.~\ref{#1}}
\def\eqref#1{Eq.~(\ref{#1})}

\def\tabref#1{Table~\ref{#1}}

\def\tabref#1{Table~\ref{#1}}

\algnewcommand\INPUT{\item[\textbf{Input:}]}%
\algnewcommand\OUTPUT{\item[\textbf{Output:}]}%

\usepackage[acronym,smallcaps]{glossaries} 
\makeglossaries 
\glsdisablehyper
\newacronym{MRF}{mrf}{Markov Random Field}
\newacronym{SGD}{sgd}{Stochastic Gradient Descent}
\newacronym{GD}{gd}{Gradient Descent}
\newacronym{PGD}{pgd}{Projected Gradient Descent}
\newacronym{ICM}{icm}{Iterative Conditional Modes}
\newacronym{DNN}{dnn}{Deep Neural Networks}
\newacronym{NN}{nn}{Neural Network}
\newacronym{PMF}{pmf}{Proximal Mean-Field}
\newacronym{PICM}{picm}{Proximal Iterative Conditional Modes}
\newacronym{BC}{bc}{BinaryConnect}
\newacronym{BWN}{bwn}{Binary Weight Network}
\newacronym{IP}{ip}{Integer Programming}
\newacronym{fc}{fc}{fully-connected}
\newacronym{REF}{ref}{Reference Network}
\newacronym{KL}{kl}{KL}
\newacronym{S}{s}{S}
\newacronym{LR}{lr}{LR}
\newacronym{KKT}{kkt}{KKT}
\newacronym{MAP}{map}{Maximum a Posteriori}
\newacronym{MDA}{mda}{Mirror Descent Algorithm}
\newacronym{MD}{md}{Mirror Descent}
\newacronym{BOP}{bop}{Binary Optimizer}
\newacronym{EDA}{eda}{Entropic Descent Algorithm}
\newacronym{ED}{ed}{Entropic Descent}
\newacronym{EGD}{egd}{Exponentiated Gradient Descent}
\newacronym{PQ}{pq}{ProxQuant}
\newacronym{STE}{ste}{Straight Through Estimator}
\newacronym{HGD}{hgd}{Hybrid Gradient Descent}
\newacronym{PQL}{pql}{PQL}
\newacronym{ELQ}{elq}{Explicit Loss-error-aware Quantization}
\newacronym{ADMM}{admm}{Alternating Direction Method of Multipliers}
\newacronym{QN}{qn}{Quantization Networks}
\newacronym{BR}{br}{BinaryRelax}
\newacronym{FC}{fc}{FC}
\newacronym{BN}{bn}{BN}
\newacronym{MLP}{mlp}{multi-layer perceptron}
\newacronym{SSL}{ssl}{self-supervised learning}
\newacronym{KNN}{knn}{KNN}

\newcommand{\sresnet}[1]{{\small ResNet#1}}

\newcommand{\vresnet}[1]{{ResNet#1}}

\newcommand{\cifar}[1]{{\small CIFAR#1}}
\newcommand{\stl}[1]{{\small STL#1}}

\newcommand{\tinyimagenet}{{TinyImageNet}}
\newcommand{\imagenet}{{ImageNet}}

\newcommand{\simclr}{{SimCLR}}
\newcommand{\simsiam}{{SimSiam}}
\newcommand{\infonce}{{InfoNCE}}
\newcommand{\directclr}{{DirectCLR}}
\newcommand{\byol}{{BYOL}}
\newcommand{\relu}{{ReLU}}
\newcommand{\tsne}{{t-SNE}}
\newcommand{\rank}{\operatorname{rank}}
\newcommand{\sgd}{{\acrshort{SGD}}}
\newcommand{\adam}{{Adam}}
\newcommand{\knn}{{\acrshort{KNN}}}
\newcommand{\comp}[2]{#1\!\circ\!#2}

\usepackage{pifont}

\usepackage{wrapfig}

\title{Understanding and Improving the Role of Projection Head in Self-Supervised Learning}

\author{%
  Kartik Gupta$^\dagger$\thanks{Part of this work was done when KG was at Amazon, Adelaide.}, 
  Thalaiyasingam Ajanthan$^{\dagger\ddagger}$, 
  Anton van den Hengel$^\ddagger$, 
  Stephen Gould$^{\dagger\ddagger}$ 
\\
$^\dagger$ Australian National University, $^\ddagger$ Amazon\\
$^\dagger$\texttt{kartik.gupta@anu.edu.au}\\
}

\begin{document}

\maketitle

\begin{abstract}
Self-supervised learning (\acrshort{SSL}) aims to produce useful feature representations without access to any human-labeled data annotations. 
Due to the success of recent \acrshort{SSL} methods based on contrastive learning, such as \simclr{}, this problem has gained popularity. Most current contrastive learning approaches append a parametrized projection head to the end of some backbone network to optimize the \infonce{} objective and then discard the learned projection head after training. This raises a fundamental question: Why is a learnable projection head required if we are to discard it after training? In this work, we first perform a systematic study on the behavior of \acrshort{SSL} training focusing on the role of the projection head layers. 
By formulating the projection head as a parametric component for the \infonce{} objective rather than a part of the network, we present an alternative optimization scheme for training contrastive learning based \acrshort{SSL} frameworks. Our experimental study on multiple image classification datasets demonstrates the effectiveness of the proposed approach over alternatives in the \acrshort{SSL} literature.
\end{abstract}

\section{Introduction}
The ultimate goal of \gls{SSL} is to obtain generalizable features from the information inherent to massive amounts of unlabelled data in a task-agnostic manner. The learned features can then be used to perform various downstream tasks using only a minimal amount of supervised training and a small set of task-specific label data. 

The \gls{SSL} objective is typically addressed through contrastive learning, where the idea is to learn features that remove the effect of data augmentations applied to the input data. Here, the intuition is that data augmentations cover the style space, which is often irrelevant to the downstream tasks. As one can imagine, prior knowledge of the downstream tasks is necessary to design data augmentations, and even then, it is a challenging problem.

Nevertheless, contrastive \gls{SSL} methods are successful in many applications~\cite{chen_big_2020,chen2020improved,richemond_byol_2020}. An important architectural choice in the majority of these methods is the use of a \gls{MLP} appended to the network (i.e., projection head) to project the backbone features into a low dimensional space before applying the contrastive loss. This projection head is discarded after training as the projected features have been found to be inferior in terms of generalization performance and the backbone features are directly used for the downstream tasks~\cite{chen2020simple}. Despite its practical importance, the role of the projection head in \gls{SSL} methods is not well-understood.

In this work, we attempt to empirically understand why a learnable projection head is required if we are to discard it after training? We would like to highlight two important observations: 
First, the projection head is a low-rank mapping. Second, the null space of the projection head is useful for generalization. 
Based on these observations, we hypothesize that {\em the projection head implicitly learns to choose a subspace of features to apply the contrastive loss}. This subspace selection addresses the shortcomings of the contrastive loss (\eg, sub-optimal data augmentations), however, this property emerges as a side-effect. In particular, the implicit subspace selection enables the projection head output to minimize the contrastive loss, while the backbone has the flexibility to learn generalizable features.

Based on this hypothesis, we argue that the data-dependent subspace selection should be considered as part of the \gls{SSL} loss function and this behavior should be enforced rather than relied upon. To this end, we formulate self-supervised learning as a bilevel optimization problem. Here, at each training step, the inner-optimization selects the best subspace to apply the contrastive loss (by optimizing the projection head) and the outer-optimization performs gradient descent on the backbone network using this updated subspace.

We perform several experiments on \cifar{-10}, \stl{-10}, \tinyimagenet{}, and \imagenet{} datasets with \simclr{}~\cite{chen2020simple} and \simsiam{}~\cite{chen2020exploring} to understand the role of the projection head. Later, we evaluate our modified optimization scheme for \gls{SSL} on \cifar{-10}, \cifar{-100} and \tinyimagenet{} datasets with the \simclr{} method. Our results, obtained by viewing the projection head as part of the loss rather than the network, show better generalization over \simclr{} validates the hypothesis on the role of the projection head.

\section{Preliminaries and Notations}
\label{sec:prelim}

Given a dataset $\gD=\{\rvx_i\}_{i=1}^n$ and a set of data augmentations $\gT$, the \gls{SSL} learning problem can be written as:

\begin{align}\label{eq:ssl}
    \min_{f,g} L(\comp{g}{f}; \gD, \gT) &\triangleq \mathop{\mathbb{E}}_{\substack{\rvx, \gB_\rvx \sim \gD\\t_1, t_2 \sim \gT}}\left[ \ell(\rvz^1, \rvz^2; \gZ_\rvx) \right]\ ,\\\nonumber
    \mbox{where}\quad \rvz^j &= \comp{g}{f}(t_j(\rvx))\ ,\ \text{for}\ j\in\{1,2\}\ ,\\\nonumber
    \gZ_\rvx &= \left\{\comp{g}{f}(t_j(\rvy))\mid \rvy\in\gB_\rvx, j \in\{1, 2\}\right\}\ .
\end{align}
Here, $g$ denotes the projection head, $f$ is the (feature) encoder, $\gB_\rvx$ is a mini-batch sampled from $\gD$ that does not include $\rvx$, and $\ell$ is the example-wise contrastive loss function. 
While there are various contrastive loss functions exist~\cite{liu2021self}, \infonce{}~\cite{chen2020simple} is the most popular and we provide its exact form below:
\begin{equation}\label{eq:infonce}
    \ell(\rvz^1, \rvz^2; \gZ_\rvx) = -\log\left(\frac{\exp(\cos(\rvz^1, \rvz^2)/\tau)}{\sum_{\rvz \in \gZ_\rvx}\exp(\cos(\rvz^1, \rvz)/\tau)}\right)\ ,
\end{equation}
where $\cos(\rvu, \rvv) = \rvu^T\rvv/\norm{\rvu}\norm{\rvv}$ is the cosine similarity, and $\tau > 0$ denotes the temperature.
Note that the loss encourages the embeddings of two augmented views of the same training example to be closer while pushing embeddings of other training examples apart.

In practice, $f$ and $g$ in~\eqref{eq:ssl} are parametrized using neural networks and $\min_{g,f}$ denotes minimizing the parameters of the respective neural networks. In this setting, the encoder $f$ is sometimes referred to as the backbone network (\eg, \vresnet{-50}) and the projection head $g$ is usually a linear layer or a \gls{MLP}.

Let us denote the backbone features by $\rvh = f(\rvx) \in \R^m$ and the projection head output as $\rvz = g(\rvh) \in \R^d$, where $m \ge d$ in practice.
Moreover, let $\gH \subseteq \R^m$ and $\gZ \subseteq \R^d$ be the vector spaces spanned by the backbone features and projection head outputs, respectively.

Once trained, the projection head $g$ is stripped away and the backbone feature $\rvh$ is directly used for the downstream tasks. 
This is motivated by the better empirical generalization performance of $\rvh$ compared to the projection head output $\rvz$ but the reasons remain elusive~\cite{chen2020simple}.

\begin{figure*}[t]
        \begin{subfigure}{0.33\textwidth}
          \includegraphics[width=\textwidth]{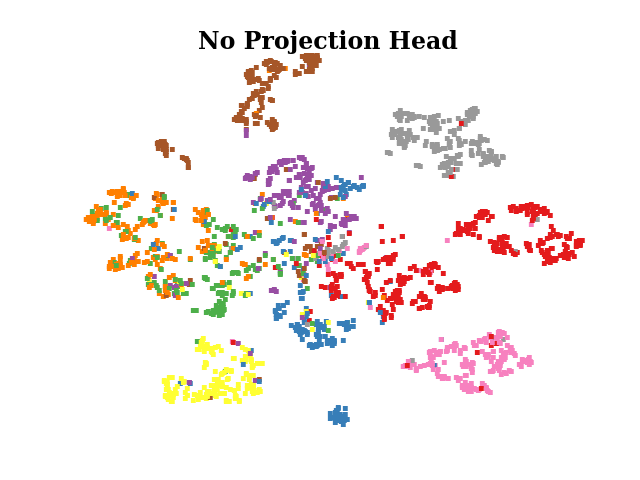}
          \vspace{-3ex}
        \caption{}          
        \end{subfigure}          
        \begin{subfigure}{0.33\textwidth}
          \includegraphics[width=\textwidth]{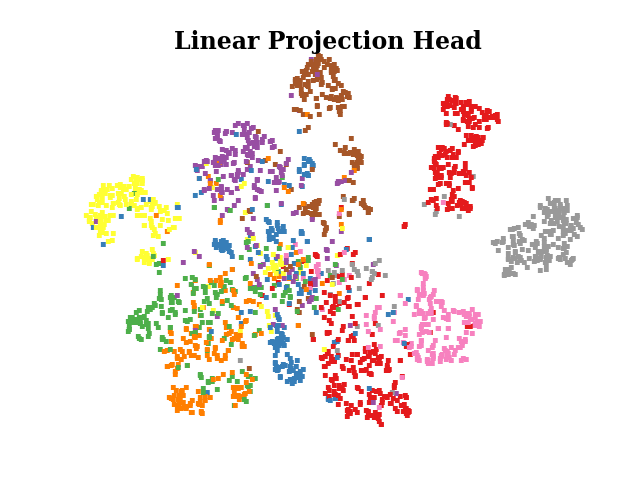}
          \vspace{-3ex}
        \caption{}          
        \end{subfigure}          
        \begin{subfigure}{0.33\textwidth}
          \includegraphics[width=\textwidth]{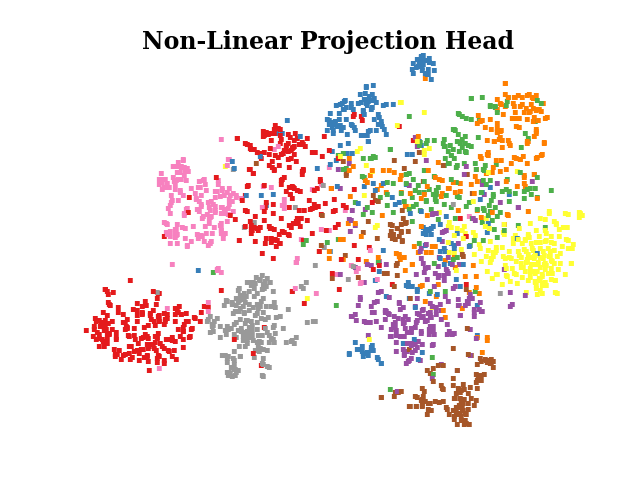}
          \vspace{-3ex}
        \caption{}          
        \end{subfigure}          
    \vspace{-1ex}
    \caption{\em \tsne{} plots for \simclr{} trained models on \cifar{-10} using \sresnet{-50} with different projection head configurations.
    The feature representations learned with no projection head are tightly clustered compared to the non-linear projection head which yields features spread across the whole space with less evidence of ``feature collapse''.
    }
\label{fig:tsne}
\end{figure*}

\begin{figure*}[t]
        \begin{subfigure}{0.33\textwidth}
          \includegraphics[height=0.7\textwidth]{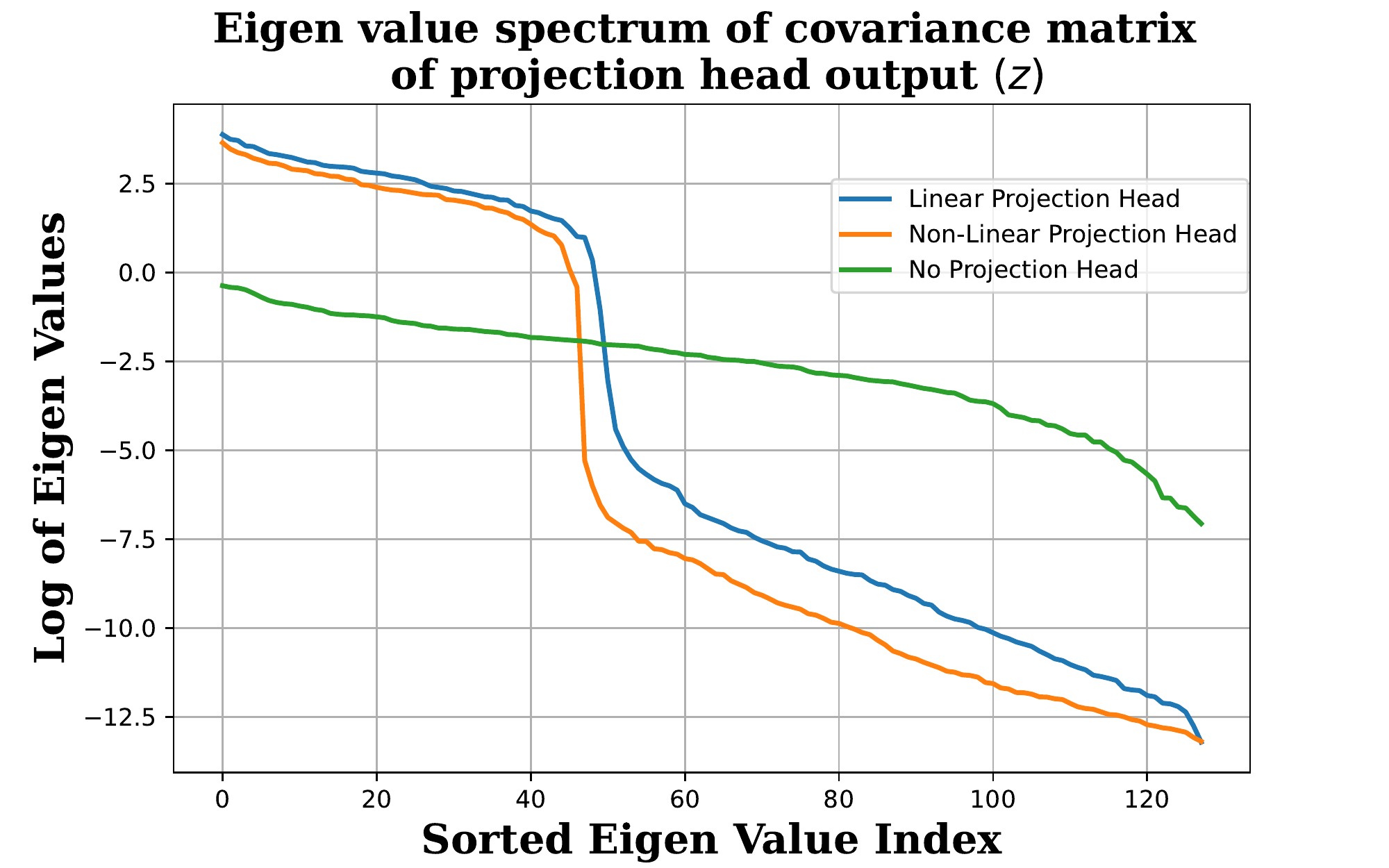}
        \caption{}          
        \end{subfigure}          
        \begin{subfigure}{0.33\textwidth}
          \includegraphics[height=0.7\textwidth]{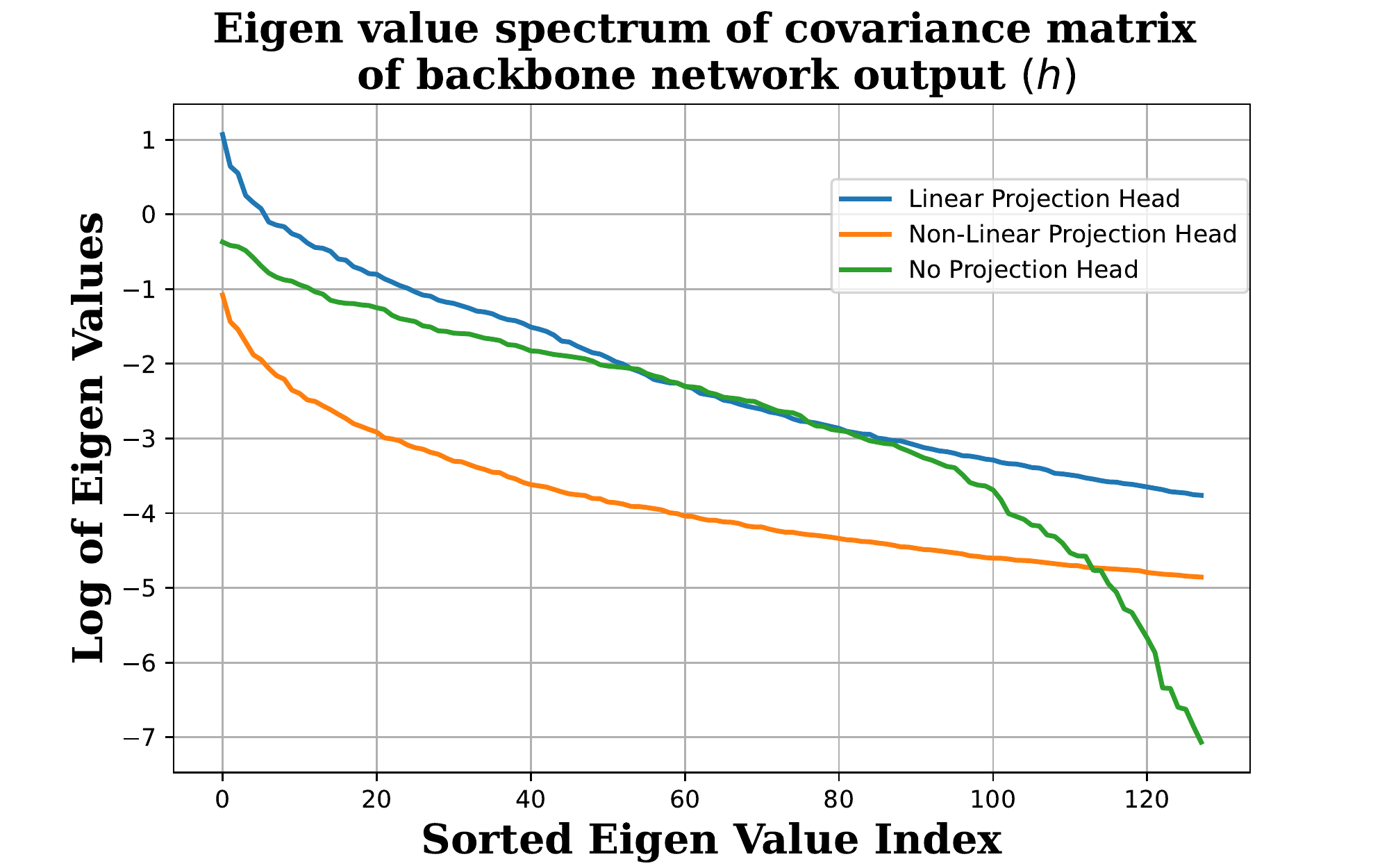}
        \caption{}
        \end{subfigure}          
        \begin{subfigure}{0.33\textwidth}
          \includegraphics[height=0.7\textwidth]{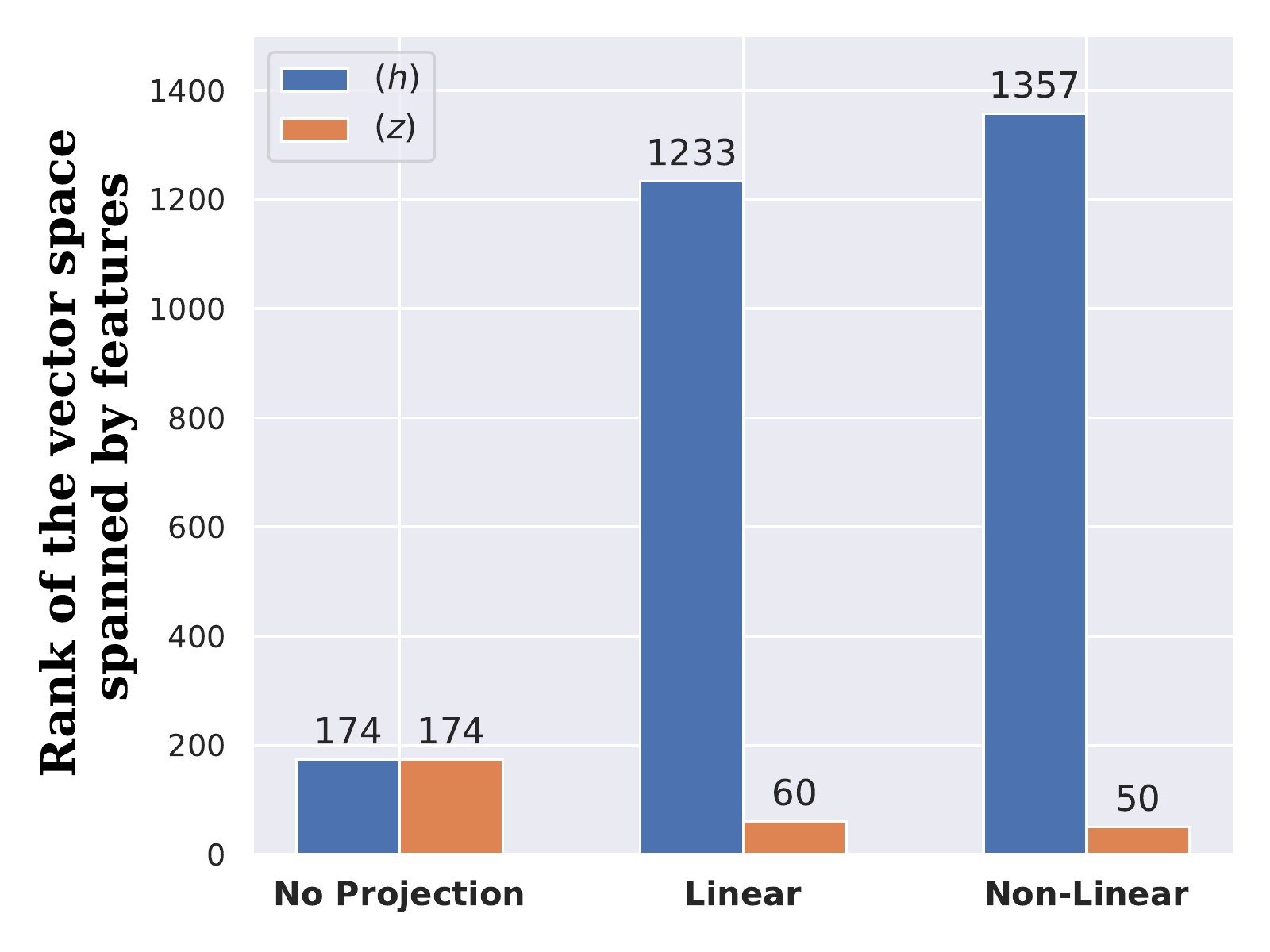}
        \caption{}
        \end{subfigure}          
    \vspace{-1ex}
    \caption{\em Eigenvalue spectrum of the covariance matrix of the (a) projection head output (\ie, the vector space $\gZ$) and (b) backbone network output (\ie, the vector space $\gH$) for different projection head configurations. For the case with no projection head, we only show top-128 eigenvalues. (c) Rank of projection head output space ($\gZ$) and backbone encoder output space ($\gH$), obtained using PyTorch ({\small \texttt{matrix\_rank}}). Note that there is a clear trend that with no projection head, to linear, to non-linear projection, the rank of $\gH$ increases and the rank of $\gZ$ decreases. 
    }
\label{fig:rank_analysis}
\end{figure*}

\section{A Closer Look at the Role of the Projection Head}
\label{sec: hypothesis}
We start by conducting an empirical study on various choices of projection heads available for contrastive learning on multiple datasets such as \cifar{-10}, \stl{-10}, \tinyimagenet{} and \imagenet{}. For this study, we use \simclr{}~\cite{chen2020simple} based self-supervised learning for all datasets except \imagenet{} for which we use \simsiam{}~\cite{chen2021exploring}.
Nevertheless, the analysis holds for other \gls{SSL} frameworks as well. 

\begin{wrapfigure}{r}{6.5cm}
        \vspace{-4.5ex}          \includegraphics[width=0.45\textwidth]{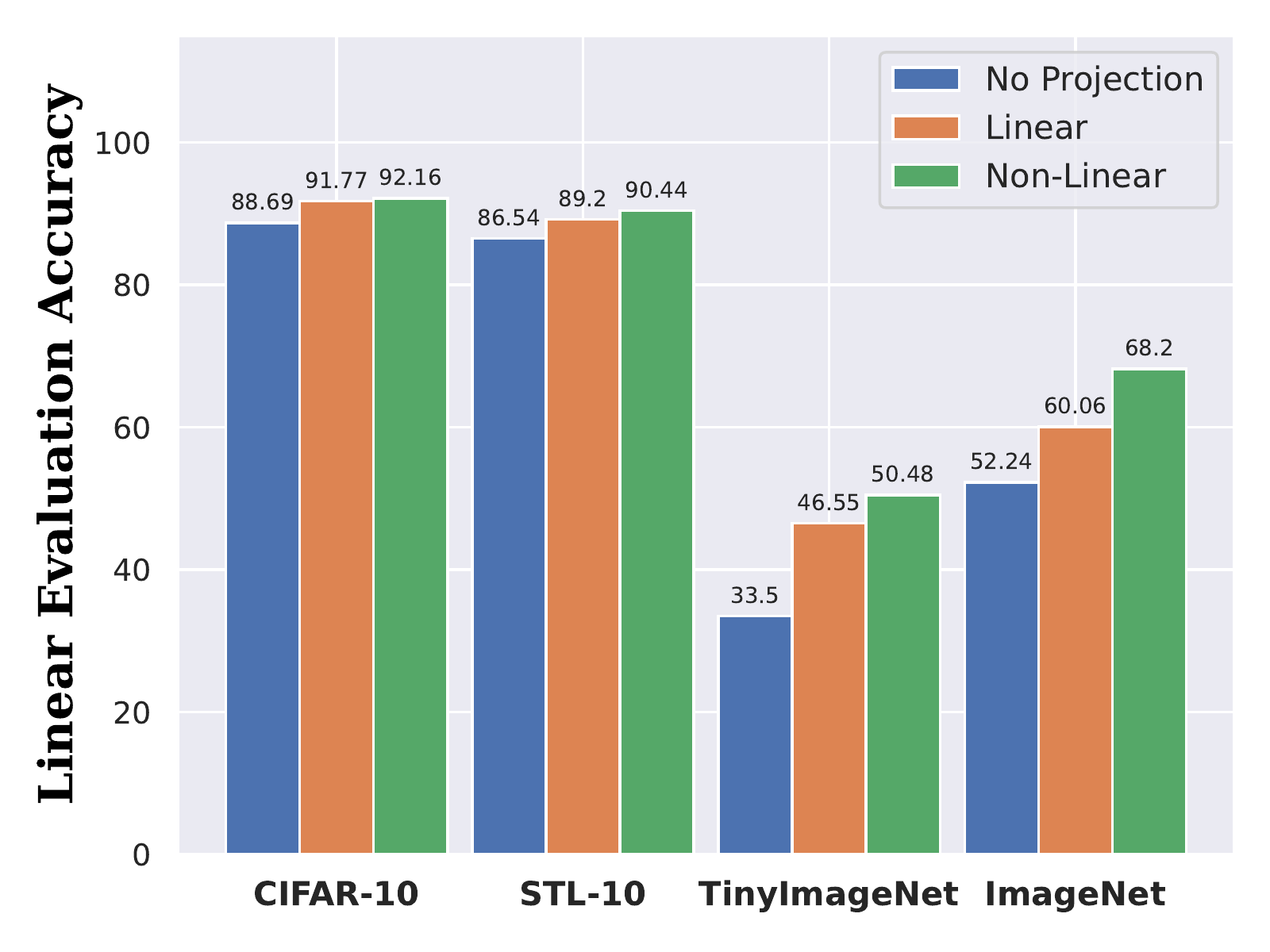}
    \caption{\em Linear evaluation accuracy of different networks with varying projection head settings, trained via \simclr{} framework on \cifar{-10}, \stl{-10}, \tinyimagenet{} datasets and \simsiam{} method on \imagenet{} dataset.}
    \vspace{-4ex}
\label{fig:acc_diffhead}
\end{wrapfigure}

For all datasets, the output dimension for projection head $d=128$, except \imagenet{} where $d=2048$. We use the \vresnet{-50} backbone with $m=2048$ and experiment with different projection heads.
We describe the different projection head configurations used for these experiments below and more details on the training regime are provided in Section~\ref{sec:exp}.

We present the linear evaluation\footnote{Linear evaluation means a task-specific linear layer is learned on top of the pre-trained \gls{SSL} features on the (small) training set of the downstream task which is then evaluated on the corresponding test set.}
results using pre-trained backbone features obtained via \acrshort{SSL} over different datasets in \figref{fig:acc_diffhead}. Observe that non-linear projection head consistently yields better performance with both \simclr{} and \simsiam{}, where the gap between no-projection head and non-linear projection head is significant. 
This has also been observed in previous works~\cite{chen2020simple}, however, the difference between linear/no-linear projection heads is not well-understood.
Specifically, this raises a non-trivial question how does the (linear/non-linear) projection head aid the training of \acrshort{SSL} objective. To this end, we analyze the rank of the learned features spaces with respect to different projection heads in the next section.

\subsection{Feature Representation with Different Projection Heads.}
To further analyse the representation quality of backbone features learned with various projection head configurations, we present \tsne{}~\cite{van2008visualizing} plots in \figref{fig:tsne} 
with no-projection, linear projection, non-linear projection heads using \simclr{} training on the \cifar{-10} dataset. Although, the feature representations learned directly on the backbone features (no projection head case) are more tightly clustered, the issue of ``feature collapse'' is easy to observe. Whereas a non-linear projection head yields features spread across the whole space with less evidence of ``feature collapse'', which is undesirable for downstream tasks. 

In order to validate this hypothesis, we further provide eigenvalue spectrum for the output of the projection head and the network backbone features in \figref{fig:rank_analysis}. This clearly shows that while all projection head variants yield low-rank backbone features, the rank of $\gH$ for the no projection head case is consistently lower than the non-linear projection head configuration.

Another consistent observation reveals that the rank of projected features ($\gZ$) is consistently lower than that of the backbone features ($\gH$).
This indicates that the contrastive learning loss tends to result in low-rank outputs $\gZ$. 

Furthermore, there is a clear correlation between the quantity $\rank(\gH) - \rank(\gZ)$ (which we call {\em rank deficit}) and the generalization performance. Specifically, the rank deficit from $\gH$ to $\gZ$ increases from no projection head ($0$), to linear ($1173$), to the non-linear projection head ($1307$) (refer to \figref{fig:rank_analysis}-(c)). The same order is apparent in the linear evaluation performance (refer to \figref{fig:acc_diffhead}).

Our observations based on the rank analysis can be summarized as follows:
\begin{align}\label{eq:rank-h-z}
    \rank(\gH) &< m\qquad\mbox{backbone features are low-rank}\ , \\\nonumber
    \rank(\gZ) &< \rank(\gH)\qquad\mbox{positive rank deficit due to $g$}\ .
\end{align}
We hypothesize that the positive rank deficit due to the projection head (\ie, $\rank(\gZ) < \rank(\gH)$) might be the main reason for $\gH$ yielding better generalization than $\gZ$. 
We conjecture that $\gZ$ becomes low-rank so as to optimize the contrastive loss, however, such a low-rank space may not be sufficiently generalizable and potentially detrimental to the downstream task performance. To understand the additional information in $\gH$ which are ignored by the projection head, we analyze the generalization performance of the null space of the projection head in the following section.
 
\subsection{Null Space Analysis for Linear Projection.}

For simplicity, we consider the linear projection head without the bias term and analyse the generalization performance of its null space (\ie, the subspace of $\gH$ which is completely ignored by $g$).

Let $\mA\in \R^{d\times m}$ be the weight matrix of the linear projection head $g$, where $m$ is the dimension of the backbone features ($\rvh$) and $d$ is projection head output dimension ($\rvz$). 
Therefore $\rvz = \mA\rvh$ and we intend to understand how $\mA$ decomposes the backbone feature space.
To this end, any vector $\rvh\in\R^m$ can be written as a sum of two orthogonal components $\rvh = \rvh_r + \rvh_n$ such that $\rvh_r \in \gR(\rmA^T)$ and $\rvh_n\in\gN(\rmA)$, where $\gR(\rmA)$ and $\gN(\rmA)$ are the range (\ie, column space) and the null space of $\rmA$, respectively.

Precisely,
\begin{align}\label{eq:null-decom}
    \rvh_r &= \mA^+\mA\rvh\  ,\\\nonumber
    \rvh_n &= \rvh - \rvh_r\ .
\end{align}
where $\mA^+$ is the right inverse, \ie, $\mA^+ = \mA^T(\mA\mA^T)^{-1}$.
Note that any vector in the null space maps to $\bfzero$ and therefore would be ignored when using $\rvz$ for the downstream tasks. This means, there exists many $\rvh\in\R^m$ corresponds to a single $\rvz$, where $\mA(\rvh - \rvh_n) = \rvz$ for all $\rvh_n\in\gN(\mA)$.
Nevertheless, to analyse the null space, we use the above decomposition~\eqref{eq:null-decom} to obtain the component corresponding to the null space for each feature vector $\rvh$.

We then evaluate the performance of different components $\rvh_{r}$, $\rvh_{n}$, $\rvh$ and $\rvz$ over pre-trained \gls{SSL} models on different datasets (\simsiam{} for \imagenet{} and \simclr{} for all other datasets) in \figref{fig:nullspace}. 
While the full backbone feature $\rvh$ is the best performing, the null space $\rvh_n$ is competitive and except in \imagenet, it outperforms the projection head output $\rvz$. Even in \imagenet{}, the null space is significantly better than the random classification. This clearly shows that the null space of the projection head has useful information for generalization despite not having any direct influence on the contrastive loss.

\subsection{Observations on the Role of the Projection Head}\label{sec:hypo}

\begin{wrapfigure}{r}{6.5cm}
        \vspace{-4.5ex}      \includegraphics[width=0.45\textwidth]{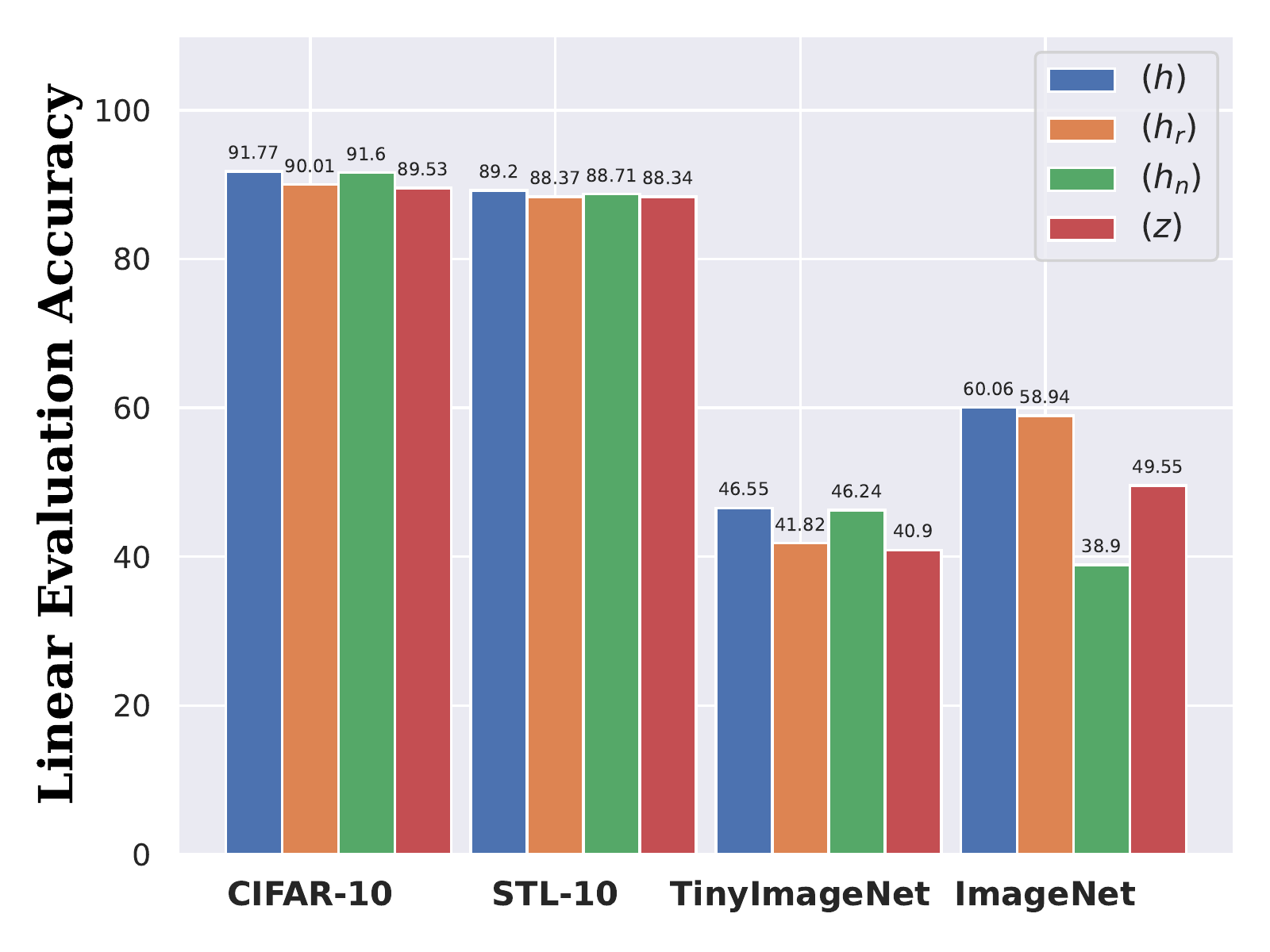}
    \caption{\em Linear evaluation accuracy on different datasets using various components of the features. While the full backbone feature $\rvh$ is the best performing, the null space $\rvh_n$ is competitive and except in \imagenet, it outperforms the projection head output $\rvz$. Even in \imagenet{}, the null space is significantly better than the random classification. This clearly shows that the null space of the projection head has useful information for generalization.}
\label{fig:nullspace}
        \vspace{-9.5ex}
\end{wrapfigure}

From the above analysis, we want to highlight the following observations: 
\begin{itemize}[leftmargin=*]
    \item The null space of the projection head is generalizable\footnote{We measure generalization using linear evaluation on the downstream task.} and sometimes it performs better than the projection head output $\rvz$. Refer to \figref{fig:nullspace}.
    \item There is a clear positive correlation between the rank deficit from $\gH$ to $\gZ$ (\ie, $\rank(\gH) - \rank(\gZ)$) and the generalization performance of the backbone features. Refer to \twofigref{fig:acc_diffhead}{fig:rank_analysis}.
\end{itemize}

Based on these observations we attempt to answer the following fundamental question: 
\begin{quote}
{\em Why is a learnable projection head required if we need to discard it after training?}
\end{quote}

As discussed earlier, the ultimate goal of \gls{SSL} is to learn generalizable features in a task-agnostic manner. This is a seemingly impossible task without prior knowledge of the downstream tasks. The best option that researchers have discovered is to formulate a contrastive loss, \ie, a loss that attempts to remove the effect of data augmentations that change \emph{style} but not \emph{content}~\cite{tsai2020self,von2021self}.
However, one major limitation of it is that the concept of style -- realized using data augmentations is not task-agnostic. Furthermore, even with prior knowledge of downstream tasks designing data augmentations to cover the whole style space without affecting the content information is unrealistic~\cite{chen2020intriguing}. This limits the generalizability of the learned features.

We hypothesize that {\em the learnable projection head is a way of mitigating the shortcomings of the contrastive loss}. Specifically, the projection head implicitly learns to select a subspace of the features $\gH$ and (non)-linearly map them to $\gZ$ to apply the contrastive loss. In this way, the contrastive loss is minimized on $\gZ$, which is encouraged to become style-invariant (hence, sensitive to the sub-optimal choice of data augmentations), whereas the backbone features $\gH$ are not forced to be style-invariant and as a result are can generalize better.

Interestingly, this property of the projection head automatically emerges as a side-effect of back-propagation. In this work, we intend to make use of this interpretation to improve the \gls{SSL} framework.

\section{Self-Supervised Learning with Adaptive Contrastive Loss}

As discussed in the previous section, if the projection head $g$ chooses the best subspace to apply the contrastive loss and is also stripped away after training, then we believe that $g$ should be considered as a part of the loss function rather than a part of the network.
Specifically, we argue that the \gls{SSL} objective should be treated as the best contrastive loss that can be obtained by searching over all (fixed-dimensional) subspaces of the features. As such we define the new objective for \gls{SSL} as
\begin{align}
    L^{\star}(f; \gD, \gT) \triangleq \min_g L(\comp{g}{f}; \gD, \gT)\ ,
\end{align}
allowing us to re-write the \gls{SSL} objective in~\eqref{eq:ssl} as:
\begin{align}
\label{eq:ssl-goal}
    \min_{f} L^{\star}(f; \gD, \gT) = \min_{f} \min_{g} L(\comp{g}{f}; \gD, \gT)\ .
\end{align}
Note that if we have access to the minimization oracle of $L$, then the optimization problems described by \eqref{eq:ssl} and \eqref{eq:ssl-goal} are equivalent. However, such an oracle does not exist in practice. To this end, we first derive an iterative optimization algorithm for the above formulation and then discuss the practical benefits of this novel perspective.

\subsection{Iterative Optimization}
Notice that in \eqref{eq:ssl-goal} we have separated the optimization with respect to $g$ and $f$, and the iterative updates can be written as:
\begin{align}\label{eq:ssl-alt}
    g^{k+1} &= \amin{g}\, L(\comp{g}{f^{k}}; \gD, \gT)\ ,\\\nonumber
    f^{k+1} &= f^{k} - \eta \nabla_f L(\comp{g^{k+1}}{f^k}; \gD, \gT) \ ,
\end{align}
where $k$ denotes the iteration number, $\eta >0$ is the learning rate, and $\nabla_f L$ denotes the gradient of $L$ with respect to the parameters of $f$\footnote{Here, for simplicity of notation we use $f$ (resp., $g$) to denote the encoding function (resp., projection head) as well as its parameters. The meaning should be clear from the context where the notation is used.}.
This bilevel optimization is computationally expensive as the inner-optimization (with respect to $g$) needs to be solved for every iteration of the outer optimization (with respect to $f$).

Therefore, for computational efficiency, we resort to a stochastic optimization strategy where for each iteration $k$ a mini-batch of data $\gD^k \subset \gD$ is used instead of the whole dataset. However, this could lead to $g$ quickly overfitting to the small mini-batch, leading to sub-optimal results. To alleviate such overfitting and allow stability, we resort to a truncated optimization with a proximal term by fixing the number of iterations to the inner optimization and use momentum based gradients.
Such an approximation is common in deep learning~(\eg,~\cite{finn2017model}).
Our final iterative updates can be written as:
\begin{align}\label{eq:ssl-alt2}
    g^{k+1} &\approx \amin{g}\, L(\comp{g}{f^{k}}; \gD^k, \gT)     + \lambda \norm{g - g^k}_2^2 
&\mbox{$l$ steps of \sgd{}},\nonumber\\
    f^{k+1} &= f^{k} - \eta \nabla_{\!f} L(\comp{g^{k+1}}{f^k}; \gD^k, \gT)&\mbox{$1$ step \sgd{}},
\end{align}
where any standard stochastic gradient descent (\sgd{}) algorithm can be used.
Here, $\lambda >0$ is the strength of the proximal term and $\norm{\cdot}_2^2$ denotes the $L_2$ norm.

Note that the optimization with respect to $g$ can be performed quickly as it is a small \gls{MLP}, and run for a small number of iterations (typically we set $l\le 10$ in our experiments). Then, the updated $g^{k+1}$ is used to compute the gradient with respect to the backbone network parameters. Overall, the computational complexity of one iteration of our algorithm is the same as the standard back-propagation.

\subsection{Practical Benefits}
According to~\eqref{eq:ssl-alt2}, it is clear that at every iteration of our algorithm, we first update the loss parameters $g$ (via inner-optimization) and use the updated loss to perform gradient descent on $f$. In this way, the most recent update on $g$ (\ie, $g^{k+1}$) is immediately propagated to $f$. This is in contrast to the standard back-propagation where the gradient of $f$ is computed with the old $g^k$. 

Therefore, based on our hypothesis in~\secref{sec:hypo}, our approach modifies the loss to improve its generalizability (\ie, selects the best subspace to apply the contrastive loss for each mini-batch) and takes a gradient step on the updated loss. 
We claim that if our hypothesis is true, then our approach should yield better generalizable features compared to the standard method.

Furthermore, despite the stochastic nature, our scheme is stable due to the truncated optimization with momentum based gradients. It ensures that the solution $g^{k+1}$ is not too far away from $g^{k}$ and the information over the full dataset is retained throughout the optimization.

\section{Experiments}
\label{sec:exp}
\paragraph{Datasets and Networks.} In order to showcase the better trainability via the proposed method, we perform experiments on image classification datasets such as \cifar{-10}, composed of $32 \times 32$ images with 10 and 100 classes, respectively, and \tinyimagenet{} \cite{tinyin}, a reduced version of ImageNet, composed of 200 classes with images resized to size $64 \times 64$, consisting of 100K training images and 10K testing image.

We use \sresnet{-50}~\cite{he2016deep} network architecture to evaluate the \acrshort{SSL} frameworks. 
\paragraph{Training and Evaluation Hyperparameters.}
For all datasets, the output dimension for projection head $d=128$, except \imagenet{} where $d=2048$. We use the \vresnet{-50} backbone with $m=2048$ and experiment with different projection heads.
We describe the different projection head configurations used for these experiments below. 
\begin{itemize}[leftmargin=*]
    \item \textit{No Projection Head}: The projection head is simply removed from the original training framework and the loss is directly optimized on the backbone features $\rvh$. 
    \item \textit{Linear Projection Head}: A linear layer with bias term is used for projecting backbone features into a lower-dimensional subspace.
    \item \textit{Non-Linear Projection Head}: A 2-layer \gls{MLP} with \relu{} non-linearity and batch normalization is used. The hidden layer dimension is $512$ for all the datasets except \imagenet{} where it is $2048$.
\end{itemize}
For both the datasets, we use the \adam{} optimizer~\cite{kingma2014adam}. We compared our method with arguably the most popular current \acrshort{SSL} method, \simclr{}~\cite{chen2020simple}. For our proposed optimization scheme, we use the same learning rate with five inner optimization steps. All the experiments are performed using \infonce{} loss with a temperature scale value of $0.5$. We use 500 epochs with learning rate $10^{-3}$ and weight decay $10^{-6}$ for both \cifar{-10} and \tinyimagenet{}. We use a mini-batch size of $256$ images for \cifar{-10} dataset and $512$ images for \tinyimagenet{} dataset. All the experimental comparisons are performed with non-linear projection head or linear projection head. The dimension of the hidden layer of the non-linear projection head $g$ is $512$. The output of the embedding size is $128$ for both \cifar{-10/100} and \tinyimagenet{} dataset. All the experiments are performed using NVIDIA Tesla V100 GPUs.

For the purpose of evaluating the generalization capabilities of pre-trained features learned using \acrshort{SSL}, we employ \knn{} based evaluation with 200 neighbours and linear evaluation where a linear layer is trained on pre-trained backbone features. For linear evaluation, we train the linear layer for 200 epochs using \adam{} optimizer with learning rate $10^{-3}$ and weight decay $10^{-6}$ for both \cifar{-10/100} and \tinyimagenet{}.

\paragraph{Augmentations used.}
Similar to \simclr{}, to generate the augmented pairs we extract crops with a random size from $0.2$ to $1.0$ of the original area and a random aspect ratio from $0.75$ to $1.33$ of the original aspect ratio for \tinyimagenet{}. For \cifar{-10/100}, we randomly extract the crops of size $32 \times 32$. We apply grayscaling to the samples with the probability $0.2$ for \cifar{-10/100} dataset and $0.1$ for \tinyimagenet{} dataset. We use color jittering with brightness, contrast, saturation, and hue configuration of ($0.4, 0.4, 0.4, 0.1$) with probability $0.8$. We also apply horizontal flipping to the image pairs with $0.5$ probability.

\subsection{Comparison against Baslines}
\SKIP{
\begin{table}[]
\centering
\begin{tabular}{l|cc|cc}
\toprule
\multirow{2}{*}{\textbf{Methods}} & \multicolumn{2}{c}{\textbf{\cifar{-10}}} & \multicolumn{2}{c}{\textbf{\tinyimagenet{}}} \\
\cmidrule{2-5}
                         & Linear         & \knn{}         & Linear           & \knn{}           \\
                         \midrule
\simclr{}~\cite{chen2020simple}                   & 91.24               &         87.54    &          49.72        &   37.89            \\
\simclr{}~\cite{chen2020simple} + Ours            &   \textbf{91.77}             &        \textbf{88.61}     &    \textbf{49.96}              &   \textbf{39.65}  \\         
\bottomrule
\end{tabular}
\caption{\em Experimental comparisons of our method against \simclr{} on \cifar{-10} and \tinyimagenet{} datasets. Our method consistently outperforms the standard training procedure of \simclr{} with both linear evaluation as well as \knn{} evaluation.}
\label{tab:comp_simclr}
\end{table}
}

\begin{table}[]
\centering
\begin{tabular}{l|cc|cc|cc}
\toprule
\multirow{2}{*}{\textbf{Methods}}                                                      & \multicolumn{2}{c}{\textbf{CIFAR-10}}     & \multicolumn{2}{c}{\textbf{CIFAR-100}}                          & \multicolumn{2}{c}{\textbf{TinyImageNet}} \\
\cmidrule{2-7}
                    & \knn{} & Linear & \knn{}   & Linear & \knn{} & Linear \\
\midrule

No Projection                       & 85.69                     & 88.07                       & 47.67                     & 54.78                       &       28.20                  &  35.04                    \\
\simclr{}~\cite{chen2020simple}                           & 87.43                     & 91.11                       & 56.10                     & 68.01                       &    38.44                     &            50.64          \\
Random Proj.                       & 86.57 & 90.75   & 51.81 & 63.25   &  32.88   & 46.40  \\
\directclr{}~\cite{jing2021understanding}             & 86.93                     & 90.19                       & 52.23                     & 65.20                       &   33.08                      &           46.71           \\
Ours                    &    \textbf{88.53}                     & \textbf{91.97}                       & \textbf{58.08}                     &   \textbf{69.12}                       &   \textbf{40.48}                      &  \textbf{51.72} \\                     
\bottomrule
\end{tabular}
\vspace{1ex}
\caption{\em Comparisons against \simclr{} on \cifar{-10/100} and \tinyimagenet{} datasets using \sresnet{-50} architecture. 
Our method consistently performs better than the standard training procedure of \simclr{} with both linear evaluation as well as \knn{} evaluation.
}
\label{tab:comp_simclr}
\end{table}

As a proof of concept, we provide the experimental evaluations of our proposed optimization scheme against the standard training regime of \simclr{} framework on \cifar{-10/100}, and \tinyimagenet{} datasets in \tabref{tab:comp_simclr}. As explained above, we use both linear evaluation and \knn{} based evaluation to show these comparisons. Our approach achieves better generalization performance using both these evaluations consistently on \cifar{-10/100} and \tinyimagenet{} datasets.
Infact our method reaches near optimum linear evaluation accuracy at just 500 epochs, which the standard \simclr{} training procedure can achieve after nearly double the training cost. This empirically demonstrates the capability of faster convergence of our alternative optimization scheme. The performance gain of our method over the \simclr{} trained using non-linear projection head is especially significant ($\approx 2\%$ on \cifar{100} and \tinyimagenet{} dataset) in case of \knn{} evaluation.

We also compare our method against alternative mechanism to a standard approach of performing contrastive learning using non-linear projection head in \tabref{tab:comp_simclr}. Our method consistently performs better than recently proposed method namely \directclr{}~\cite{jing2021understanding} as well alternative approach of performing \simclr{} with a fixed randomly initialized projection head. 

\subsection{Analysis on Different Procedures to Obtain the Projection Head}
We now provide an experimental comparison on different possible fixed projection heads that can be used to replace a trainable projection head to further validate that a trainable projection head indeed aids in better training of \simclr{} training procedure and similar benefits cannot be achieved with a fixed projection head. In this study, we resort to \knn{} evaluation for comparative analysis and linear projection heads and employ \sresnet{-50} architecture.
Our analysis on different projection heads can be divided into \textit{fixed projection head} and \textit{moving projection head}.
We now discuss the multiple variants of different projection head schemes that we use for this study.
\paragraph{Fixed Projection Head.} In this setting, we keep the projection head parameters fixed throughout the training procedure of \simclr{}. The fixed projection head can be obtained in the following variations:
\begin{itemize}[leftmargin=*]
    \item \textit{Random Initialization:} A linear projection head obtained via standard network initialization~\cite{glorot2010understanding} is employed in this case.
    \item \textit{\simclr{} based pre-training:} Here, we evaluate a fixed projection head obtained at the end of standard \simclr{} training. Note, the idea here is to evaluate if there exists an optimal projection head that can be used in the training procedure of \simclr{} as a fixed projection head (but possibly suboptimal with respect to the current backbone network parameters). 
    \item \textit{\directclr{}:} This is a recent concurrent work~\cite{jing2021understanding} where the proposed fixed projection head takes the form of a fixed low-rank diagonal matrix and is shown to outperform a trainable linear projection head. Their proposed fixed projection head essentially translates into \simclr{} objective onto a subset of backbone features $\rvh$.
\end{itemize}

\paragraph{Moving Projection Head.} In the standard training procedure of \simclr{} framework, projection head parameters are updated in the same backward pass as the backbone encoder. Although, as shown via our proposed approach that this objective can achieve more efficiently via an alternating optimization scheme. We now present some other alternatives for the training of the projection head which are described below.
\begin{itemize}[leftmargin=*]
    \item \textit{PCA based Projection:} Since our empirical observation demonstrates the projection head is low-rank. It is straightforward to estimate the projection head using principal components of $\gH$. Therefore, to evaluate the efficacy of principal components as projection head, we perform PCA at the end of each epoch on a subset of the training dataset and use top-$k$ or bottom-$k$ principal components as fixed projection head during the next epoch. 
    \item \textit{Slow updates on Projection:} Another alternate optimization is via optimization of $g$ in an optimization setting where $g$ is optimized separately at the end of each epoch via optimization on the subset of dataset until convergence, namely Slow-Optimal or via optimization as a single step update via accumulated gradients of $g$ during the training epoch of $f$, namely Slow-Single. 
\end{itemize}

\begin{table}
\centering
\begin{tabular}{ll|c}
\toprule
\multicolumn{2}{c|}{\textbf{Methods}}                                & \textbf{\knn{} Accuracy} \\
\midrule
\multirow{3}{*}{\textbf{Fixed}}     & No Projection & 28.20 \\
                                        & Random Init        & 32.88        \\
                                      & DirectCLR~\cite{jing2021understanding}          & 33.08        \\
                                      & Pre-trained SimCLR & 32.98        \\
\midrule
\multirow{4}{*}{\textbf{Moving}}    & PCA top-128        & 32.19        \\
                                      & PCA bottom-128     & 31.58        \\
                                      & Slow-Single        & 33.06        \\
                                      & Slow-Optimal       & 32.29        \\
\midrule
\multirow{2}{*}{\textbf{Trainable}} & SimCLR~\cite{chen2020simple}             & 38.44       \\
                                      &  Ours      & \textbf{40.48} \\       
\bottomrule
\end{tabular}
\vspace{2ex}
\caption{\em Experimental comparisons for KNN evaluation using different types of fixed, moving, and trainable linear projection head used for \simclr{} pre-training on \tinyimagenet{} dataset using \sresnet{-50} architecture. Note, both trainable versions of projection head consistently outperform all the fixed and slow moving alternatives and our proposed optimization scheme outperforms the standard \simclr{} training procedure.}
\label{tab:ablation_projn}
\end{table}

We present the experimental comparisons of \knn{} accuracies with above explained projection heads using \simclr{} objective on \tinyimagenet{} dataset in \tabref{tab:ablation_projn}. Consistent to our observations in \tabref{tab:comp_simclr}, our proposed optimization scheme performs better than standard \acrshort{SSL} optimization scheme even for linear projection head. It is also clear from these comparisons that trainable projection head outperforms various forms of slow-moving or fixed projection heads. 

In fact, our observations are contradictory to \directclr{}~\cite{jing2021understanding}, which claims a fixed linear projection head can outperform a trainable linear projection head. Though, all the fixed projection head do outperform \simclr{} trained model without projection, \directclr{} performs similar to the projection head setting where a fixed randomly initialized projection is employed. Interestingly, a pre-trained projection head obtained from \simclr{} training performs significantly worse and roughly similar to the random fixed projection head. 
This clearly indicates that there is possibly no such fixed projection head that can outperform a trainable alternative. The worse performance for fixed pre-trained projection head can be accounted to the fact that for any fixed projection head $g$, the backbone network $f$ can, in principle, be trained to achieve the same local minimum as without $g$. Thus, a fixed projection head is unable to aid in mitigating the shortcomings of contrastive loss. 

Based on our observations, the linear projection head in most cases is low-rank. Thus, this simple implicit condition directs us towards a technically sound alternative, PCA as a replacement of the trainable projection head. Our analysis reveals that principal components computed at each epoch are worse than a fixed random initialization alternative. This further validates our hypothesis that a crucial condition on the projection head is to minimize the \infonce{} objective.

Our slow-moving version (Slow-Single) of alternate optimization indicates, that decoupling of update steps for backbone encoder and projection is beneficial and is thus able to achieve better performance than all the fixed projection head schemes. However, Slow-Optimal performs worse than Slow-Single which indicates a truncated optimization on projection head $g$ is better than training to optimality.

\vspace{-1ex}
\section{Related Work}
\vspace{-1ex}

Self-supervised learning (\acrshort{SSL}) literature has become increasingly popular in the last few years with the promise of boosting performance in various application domains where obtaining large volumes of unlabeled data is cheap, including vision and language. Due to space constraints, we briefly discuss previous works that are closely related to our work and/or inspired our thinking, and we refer the interested reader to the surveys~\cite{jing2019self,jaiswal2021survey,liu2021self} for a comprehensive study.

Here, we mainly consider example-wise contrastive learning approaches, which can be categorized into methods that require explicit negative samples~\cite{chen2020simple,chen_big_2020,he2019momentum,chen2020improved,ermolov_whitening_2020,wang_understanding_2020} and those that are negative-sample-free~\cite{grill2020bootstrap,chen2021exploring,zbontar2021barlow}.
In the former methods, for a given input sample, all other samples in the mini-batch are regarded as negatives and the loss is encouraged to pull various augmentations of the same sample together in the feature space while pushing the features from the other samples (\ie, the negatives) apart.
This has two main issues. First, the notion of negative samples is unclear without the label information; and, second, these methods require large mini-batches to compute effective statistics for training.

Different strategies have been explored to obtain better positive and negative samples to apply the contrastive loss within the \gls{SSL} framework~\cite{chuang2020debiased,dwibedi2021little}
On the other hand, the latter methods circumvent the requirement of negative samples altogether by introducing asymmetry in the network architecture~\cite{grill2020bootstrap,chen2021exploring} or by modifying the loss function~\cite{zbontar2021barlow}. 
In both types of methods, the projection head is an integral part of the architecture design and it is discarded after training. This is true even for \gls{SSL} approaches that do not use example-wise contrastive learning~\cite{caron2020swav,li2020prototypical}.

There are some efforts to theoretically understand the \gls{SSL} methods~\cite{wei2020theoretical,haochen2021provable}, the role of data augmentation~\cite{von2021self,tsai2020self}, and some empirical analyses of the contrastive loss~\cite{chen2020intriguing} and the predictor in the so-called \byol{} framework~\cite{tian_understanding_2021}. Nevertheless, to the best of our knowledge, there is no work that attempts to understand the role of the projection head in the learning process, or indeed, why generalizability improves when it is stripped away for downstream tasks. 

Closest to our work is the concurrent work \directclr{}~\cite{jing2021understanding}, which shows that the projection head becomes low-rank due to strong data augmentations that distort the content information in the input data. 
Nevertheless, their final approach is to directly optimize the backbone features with a fixed projection head. Whereas we empirically show a trainable projection head enhances performance and introduce an improved optimization scheme for \gls{SSL}. Even though \directclr{} is a concurrent work, we compare it in our experiments and demonstrate that while a fixed projection yields better generalization in the early stages of training, a learnable projection head eventually outperforms it.

\section{Limitations}
\label{sec:AppendixLimitations}

This work is the first of a kind study on the role of the projection head in the \acrshort{SSL} training and thus it has certain limitations and open questions for future research.
Although contrastive learning-based \acrshort{SSL} has made significant progress in the pre-training domain, the underlying procedure remains more so as a black-box system. Some recent attempts~\cite{wei2020theoretical,haochen2021provable} have been made in this direction but we believe there is still a gap between the theoretical understanding of contrastive learning-based \acrshort{SSL} and its empirical success. 

Our work mainly attempts to understand the behavior of \acrshort{SSL} training by means of extensive empirical analysis, though the theoretical understanding of the projection head remains part of the aforementioned future work. Even though our proposed alternative optimization scheme has been shown to yield better generalization capabilities, we do not provide any theoretical guarantees for the success of our method in this paper. The proposed method aims to improve the trainability of the projection head and in turn \acrshort{SSL} training procedure.

Our observations are consistent on multiple small-scale or large-scale datasets but the alternative optimization scheme has been considered only on small-scale datasets namely \cifar{-10}, \cifar{-100} and \tinyimagenet{}. Nevertheless, the observations and analysis that we presented in this paper provide a foundation for further work, and we believe they are valuable to other researchers in the field.

\section{Conclusion}

In this work, we performed an extensive empirical study to understand the role of the projection head commonly used in the training procedure of recently popular contrastive based \acrshort{SSL} methods. We have made an attempt to answer an important and overlooked question in the literature: \textit{Why is a trainable projection head used in \acrshort{SSL} training and then discarded during the evaluation?} By performing various experimental analyses, we have shown that a trainable projection head cannot be replaced with a fixed projection head in the standard \acrshort{SSL} training procedure. Based on our observation, we also present a new objective for \acrshort{SSL} that aims to consider the projection head as a part of the loss function and optimize this new objective function via an alternating optimization.

\bibliography{ssl}
\bibliographystyle{plain}


\end{document}